\documentclass[nohyperref]{article}

\usepackage{microtype}
\usepackage{graphicx}
\usepackage{subfigure}
\usepackage{booktabs} 

\usepackage{hyperref}

\usepackage[accepted]{icml2023}

\usepackage{amsmath}
\usepackage{amssymb}
\usepackage{mathtools}
\usepackage{amsthm}

\usepackage[capitalize,noabbrev]{cleveref}

\theoremstyle{plain}

\theoremstyle{definition}

\theoremstyle{remark}

\usepackage[textsize=tiny]{todonotes}

\icmltitlerunning{Topological Data Analysis Guided Zero Shot Segmentation in Biological Imaging}

\begin{document}

\twocolumn[
\icmltitle{Topological Data Analysis Guided Segment Anything Model Prompt Optimization for Zero-Shot Segmentation in Biological Imaging}


\icmlsetsymbol{equal}{*}

\begin{icmlauthorlist}
\icmlauthor{Ruben Glatt}{llnl,equal}
\icmlauthor{Liu Shusen}{llnl,equal}

\end{icmlauthorlist}

\icmlaffiliation{llnl}{Lawrence Livermore National Laboratory, Livermore, CA, USA}

\icmlcorrespondingauthor{Ruben Glatt}{glatt1@llnl.gov}

\icmlkeywords{Machine Learning, SAM, Prompt Optimization, TDA}

\vskip 0.3in
]



\printAffiliationsAndNotice{\icmlEqualContribution} 

\begin{abstract}

Emerging foundation models in machine learning are models trained on vast amounts of data that have been shown to generalize well to new tasks.
Often these models can be prompted with multi-modal inputs that range from natural language descriptions over images to point clouds.
In this paper, we propose topological data analysis (TDA) guided prompt optimization for the Segment Anything Model (SAM) and show preliminary results in the biological image segmentation domain.
Our approach replaces the standard grid search approach that is used in the original implementation and finds point locations based on their topological significance.
Our results show that the TDA optimized point cloud is much better suited for finding small objects and massively reduces computational complexity despite the extra step in scenarios which require many segmentations.

\end{abstract}

\section{Introduction}
\label{sec:introduction}


In recent years, advances in machine learning research and data availability have led to a new class of knowledge preservation in general-purpose models, termed \textit{foundation models} \cite{bommasani2021opportunities}.
These models are powerful neural networks that have been trained on extremely large datasets and usually generalize well to unseen data.
Initially, foundation models were focused on natural language processing such as BERT \cite{devlin2018bert} and GPT \cite{radford2018improving}.
However, in recent years work has expanded to other applications like image synthesis \cite{dosovitskiy2020image}, protein sequencing\cite{rives2021biological}, and others.

With the recent introduction of the Segment Anything model (SAM) \cite{kirillov2023segment}, the image segmentation task has also entered the era of foundation models.
This foundational model introduces the combination of an image with a multi-modal segmentation prompt that describes the segmentation task.
The prompt can be a point grid, bounding boxes, an image mask, or even text.
This flexible way of injecting knowledge and context into the segmentation process supports the state-of-the-art generalization performance that the model exhibits and makes it a perfect candidate for zero-shot learning approaches.
The segmentation process remains costly, specifically in tasks with a large number of similar objects in the image that need to be categorized into individual segments.
These type of tasks are particularly prevalent in biological applications, e.g., various cells or biomass, in which pure user-guided prompt-based segmentation, i.e., user click on a segment is becoming feasible.

\begin{figure*}[ht]
\begin{center}
\centerline{\includegraphics[width=\textwidth]{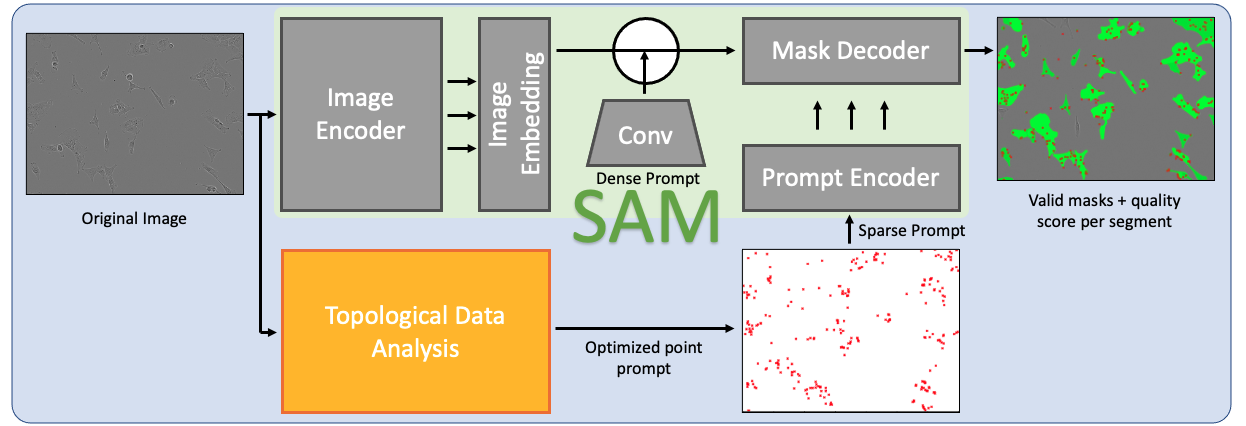}}
\caption{Architecture of the Segment Anything Model extended with topological data analysis guided prompt optimization.}
\label{fig:overview}
\end{center}
\vskip -0.2in
\end{figure*}


The SAM model already supports an automatic segmentation pipeline, however, the mechanism is rather rudimentary and with limited capability.
Without additional information, the default approach to the segmentation task is the use of an equidistant point grid, that describes the areas of the image that will be taken into consideration for the segmentation.
Such simple solutions work reasonably well for natural images as meaningful segments are often large and different segments often cover a sizable portion of the image.
However, for many biology applications, due to the large number of small objects (e.g., cells) that reside in a potentially sparse space, the grid-based search scheme for automatic segmentation meets significant challenges.
In order to achieve better coverage, we can of course increase the grid resolution, however, that will induce a quadratic increase in the number of input prompts, which translates into significantly increased computational costs.
Moreover, the grid-based approach totally ignores the information already present in the image, which can and should be leveraged to improve the models overall capability.

In this work, we introduce a novel approach to augment automatic prompt generation by adopting topological data analysis (TDA) \cite{pascucci2010topological, wasserman2018topological, liu2019scalable} in a pre-processing step to generate meaningful and efficient prompts for an image as shown in Fig.~\ref{fig:overview}.
Instead of randomly probing an image for meaningful segment locations or using a fixed point grid, TDA extracts local extrema in the image domain, which then become the natural surrogate to indicate the rough location of a potential segment.
More importantly, TDA provides a well-defined and simple-to-interpret mathematical foundation for the analysis of features in different scales, making it less likely to miss important objects and make unpredictable or potentially unreliable decisions compared to purely static learning approaches.

In this work in progress, we show the impressive results of applying the proposed TDA-based prompt optimization approach on a small self-generated synthetic dataset as a proof of concept and then show the application on selected samples of the LIVECell dataset\footnote{Available for download: https://github.com/sartorius-research/LIVECell}.
We show that the proposed method not only helps to identify a more complete segmentation but also utilizes an order of magnitude less computational budget for the segmentation task.


\section{Related Works}

The proposed work is built around the concept of prompt engineering of a segmentation foundation model, which is related to a number of topics. 

\textbf{Foundational Models} As discussed in the introduction, the early work of foundation models have primarily focus on text data \cite{devlin2018bert, liu2019roberta, radford2018improving}
However, the latest foundation models bridge the gap between disciplines and come in the form of multimodal models \cite{radford2021learning} that have been trained on different types of data, e.g. images and text, and benefit from the synergy between those datasets.
Various of these models allow \textit{prompting} techniques that are used as input to the model and describe a desired task for the model to perform, e.g. GPT-3 models \cite{brown2020language}.
Based on their generality, these models have been shown to be excellent at few-shot learning \cite{brown2020language,schick2020s,tsimpoukelli2021multimodal}.
In the text and vision space, models like CLIP \cite{radford2021learning} or ALIGN \cite{jia2021scaling} are trained to relate text and image and allow to extract descriptions from an image or the generation of an image from the text as for example in DALL-E \cite{ramesh2021zero}.

\textbf{Prompt Optimization} With the advancement of the foundation model, effective few-shot or even zero-shot generalization becomes quite feasible, however, how the problem is framed, i.e., \emph{prompt} that is given to the model, plays a crucial role in determining the effectiveness of prediction. An additional step is often required to either correct the potential bias in the model \cite{zhao2021calibrate} or optimize the prompt to circumvent the better results \cite{chen2022knowprompt}. 
However, almost all existing prompt engineering or prompt optimization works \cite{chen2022knowprompt, jiang2022prompt, liu2022design, strobelt2022interactive} are revolving around text-based model input, as it is the most common input for current foundation models.


\textbf{Topological Data Analysis}
Topological data analysis \cite{edelsbrunner2022computational} is an emerging field in which computational methods are used to study mathematical concepts originally raised in the area of topology. The process involves various discretization and approximation of the underlying theory of topology. In this work, the center of attention on the utilization of persistence homology \cite{edelsbrunner2000topological} that allows us to explore the local extrema of interests at varying degrees of detail based on their topological significance. Despite the wide adaptation of TDA in scientific domains \cite{carlsson2020topological, bennett2011feature}, it is still a relatively underutilized technique in the context of machine learning research. 

\section{Method}

The general SAM pipeline is sketched out in Fig.~\ref{fig:overview}.
It requires two inputs for the image segmentation, the original image and a prompt.
The prompt can be comprised of a multi-modal input and be any combination of a dense pixel mask, a point grid, a natural language description, or a bounding box, or multiples of them.
Here, we are focusing on the point grid prompt as objects in biological images are often hard to describe with text or there is missing information for any of the other prompts.

\begin{figure}[!htbp]
\begin{center}
\centerline{\includegraphics[width=0.8\columnwidth]{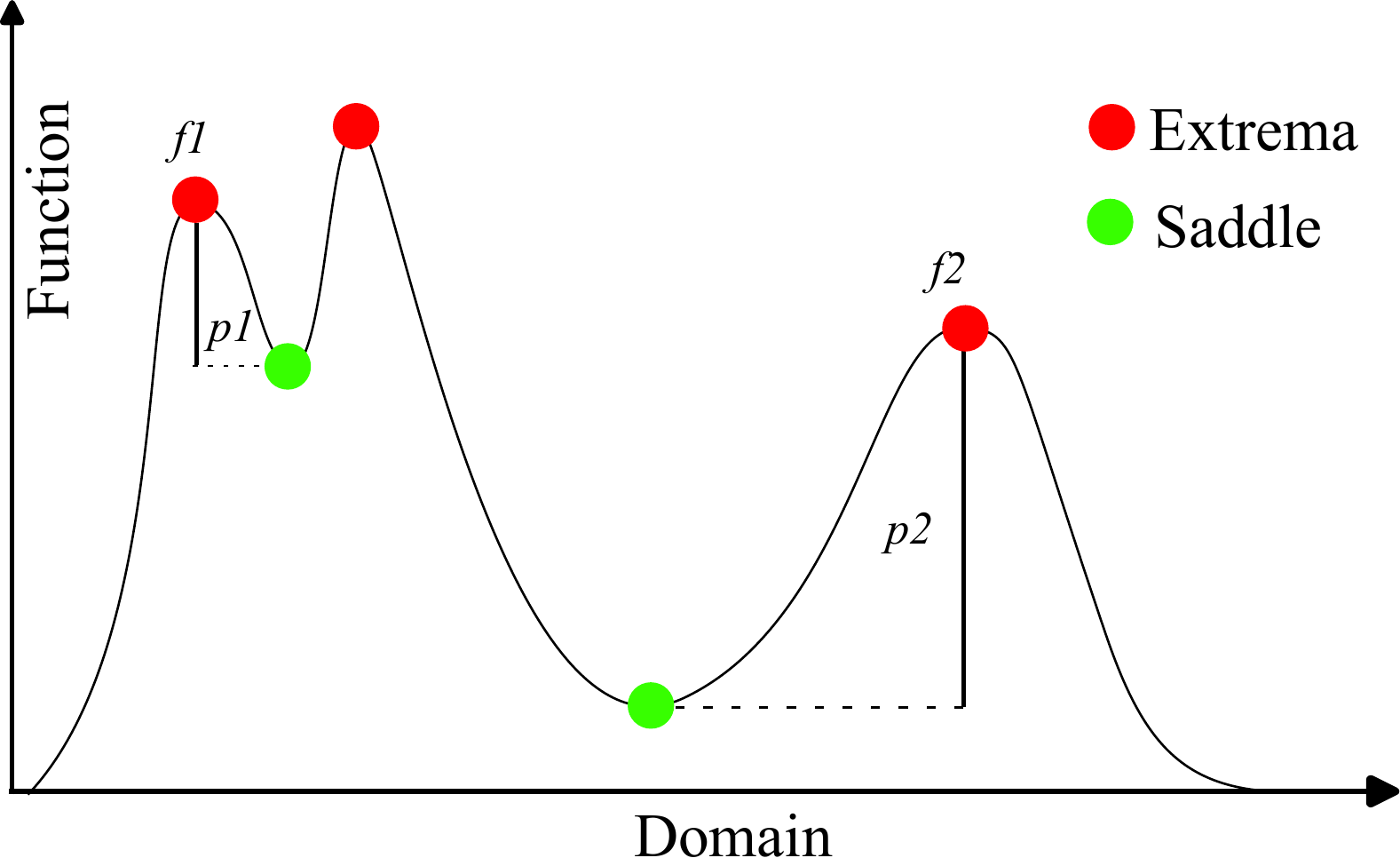}}
\caption{An illustration of topological data analysis. We can identify local extrema and saddle in the domain, the importance of a given extrema can be estimated through its persistence, which can be described as the function distance from the current extrema and the closest saddle (p1, p2 for the two maxima).}
\label{fig:topo}
\end{center}
\vskip -0.2in
\end{figure}

The key challenge to effectively use SAM for biological imaging datasets is the generation of a sensible prompt, e.g. identifying relevant points in an image.
The prompt needs to be in the general location of objects and ideally only few or a single prompt is needed for each segmentation. 
We are addressing this challenging problem of prompt optimization by finding a reduced set of points that are ordered by topological significance.
%
We propose TDA as a principled way to identify such regions of interest by examining local extrema (points of interest) and eliminating duplication through topological simplification (using the least amount of prompts for a single object).
For a given image, we can construct a function $ f: \mathbb{R}^2 \rightarrow \mathbb{R}$, where the image coordinate space $\mathbb{R}^2$ is the domain, and the value of the image is the function defined over the domain. 
Often one can see a general separation in terms of pixel value between a background and objects of interest, though this distinction alone is not enough for finding a meaningful segmentation.
To explain the concept, we rely on a simplification in 1D space as illustrated in Fig.~\ref{fig:topo}.
Simply selecting high function values would not result in meaningful separation of interesting points as it may lead to missing out of the local maximum on the right depending on a threshold.
To make sure all relevant points are captured, TDA considers regional aspects and is able to capture similar structures even though the absolute values might be different.


Persistent homology \cite{edelsbrunner2000topological, perea2018brief} and persistence-based simplification from the topology domain provide an effective tool to describe the importance of individual local extrema, which can help resolve the prompt selection challenges.
The persistent homology introduces the concept of \emph{persistence}. In the specific setting of the simple 1D function, we can understand the \emph{persistence} as the function value difference between the local extrema (red) and the nearest saddle points (green). For the extrema $f1$, the \emph{persistence} value is $p1$, and $p2$ is the \emph{persistence} for $f2$. Even though $f1 > f2$, based on \emph{persistence} $p1 < p2$, we can say $p2$ has a more significant topological feature. This structure-aware ranking of extrema forms the basis of our TDA prompting pipeline.  In this work, we utilize the Topology Toolkit \cite{ttk} to implement the extraction of local extrema and the persistence-based simplification.



\section{Experiments}
\label{sec:experiments}

In this section, we are showing the preliminary results of point prompt optimization for image segmentation tasks.
The first example here is based on a synthetic dataset that was generated to test the approach and make sure that the expected benefits are based in reality.
The second example then draws from a real world high-quality biology dataset that is openly available and provides a good benchmark for our approach.

\subsection{Synthetic data}

In the synthetic dataset we generated images with small ovals that represent cells in micro imaging scenarios.
For our experiments, we generated 10 images, each with 80 randomly distributed and sized ovals.
For each image we ran different prompt settings and recorded running time and success rate, as well as a quality
value provided by SAM.
We report the results in table~\ref{tab:results} based on the prompt setting, averaging results over the 10 images.

\begin{table*}[t]
\caption{Experimental observations as averages over 10 generated images of the synthetic dataset.}
\label{tab:results}
\vskip 0.15in
\begin{center}
\begin{small}
\begin{sc}
\begin{tabular}{lccccccr}
\toprule
Objective & Grid16 & Grid32 & Grid64 & Random256 & Random1024 & Random4096 & TDA \\
\midrule
Time[s]        & 34 & 92 & 310 & 38 & 87 & 327 & \textbf{24}\\
Accuracy[\%]   & 9.625 & 40.375 & 96.24 & 7.875 & 34.25 & 78.375 & \textbf{99.625}\\
Quality    & 0.9478 & 0.9440 & 0.9434 & \textbf{0.9544} & 0.9431 & 0.9427 & 0.9450\\
\bottomrule
\end{tabular}
\end{sc}
\end{small}
\end{center}
\vskip -0.1in
\end{table*}

Different settings for the point prompts are shown in the Appendix, Fig.~\ref{fig:synthetic_data_experiments}, where we draw the used point grid over the image and found segments.
It can be seen in this figure, that our engineered prompt finds 83 points of interest based on the TDA approach and identifies all objects in the image.
The other images show a standard grid approach with a 16x16 point grid that only detects 11 of the 80 objects, as well as a 64x64 grid that seems to be very dense but still only detects 78 out of 80, and a random point grid with 4096 points that even misses 13 out of 80.
Despite taking much less time for the segmentation task, our optimized prompt achieves a much better result, highlighting the potential gains of the approach.
TDA actually found $100\%$ of the objects but SAM failed to recognize 4 over the whole dataset.
The closest performance was achieved by the the more dense 64x64 grid but it needed $~13x$ the time to evaluate all points.

\subsection{Biology dataset}

For the first evaluation of the real-world feasibility of our prompt optimization, we considered samples from the LIVECell dataset \cite{edlund2021livecell}.
The LIVECell dataset is a large, high-quality, manually annotated and expert-validated dataset of phase-contrast images featuring a diverse set of cell morphologies and culture densities.
This dataset includes 5,239 manually annotated, expert-validated, Incucyte HD phase-contrast microscopy images with a total of 1,686,352 individual cells annotated from eight different cell types, averaging 313 cells per image, everything organized in predefined splits of training (3188), validation (539), and test (1512) sets.
The authors believe that using this diverse set of cells and confluence conditions in the LIVECell dataset can help to train deep-learning-based segmentation models more accurately, providing a robust and accurate way to train neural networks.
We believe that the introduction of foundation models for image segmentation will enable subject matter experts to generate this kind of dataset with much less manual labor and further support the efforts of building high quality datasets to train expert models.
As this is a work in progress, we did not evaluate the full dataset but randomly selected a small subset of different cell types to show the general feasibility of our approach.

\begin{figure}[!htbp]

\begin{center}
\centerline{
    \includegraphics[width=0.9\columnwidth]{A172_Phase_C7_1_00d08h00m_3}
}
\vskip 0.05in
\centerline{
    \includegraphics[width=0.9\columnwidth]{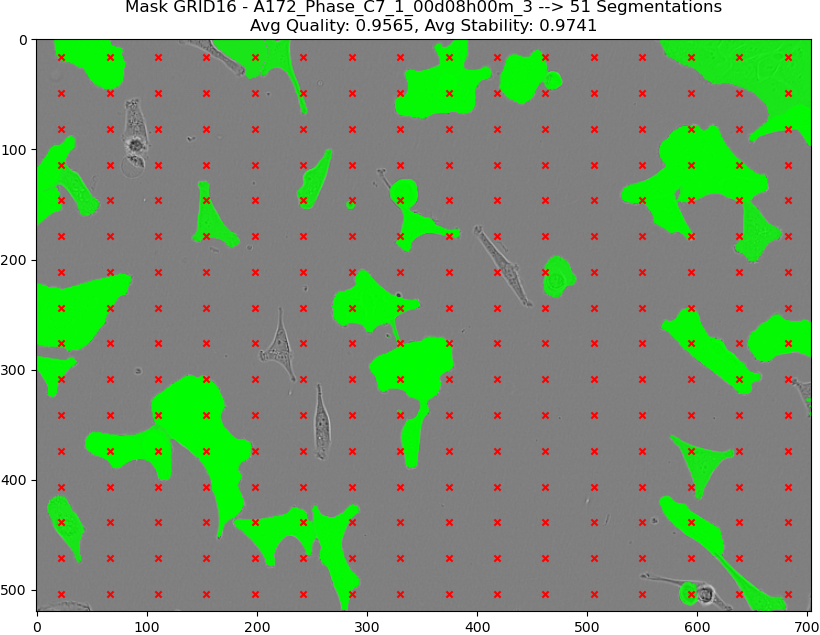}
}
\vskip 0.05in
\centerline{
    \includegraphics[width=0.9\columnwidth]{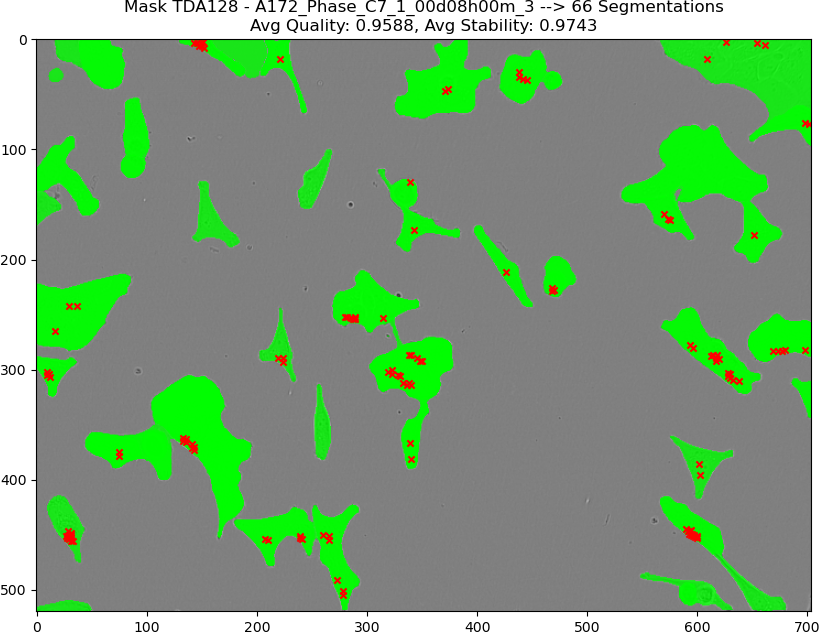}
}  
\caption{Segmentation results: $(top)$ Original image, $(middle)$ Point grid with a 16x16 raster, and $(bottom)$ TDA approach for optimized prompt with a budget of 128 points.}
\label{fig:example}
\end{center}
\vskip -0.2in
\end{figure}

Fig.~\ref{fig:example} shows the differences in the approach.
The segments in the original image are rather large compared to the image size so that most segments are captured using the standard approach with a 16x16 grid (256 points).
However, it is clearly visible that smaller objects fall through the grid and remain undetected.
The optimized prompt on the other hand is clearly concentrated on areas of interest, uses only 128 points, takes $30\%$ less time, and still clearly captures more segments, a pattern we observed in many cell images.

\section{Discussion}
\label{sec:discussion}

Our initial experiments with point grid prompt optimization have shown very promising results.
While the standard grid captures a good portion of the larger objects in the image, it struggles with objects that fall through the grid.
Using TDA guided prompt engineering can massively reduce the computational efforts for the segmentation task while generally achieving better results. 
In future research, we plan to further improve our method for prompt optimization and integrate it as a support tool for annotating biology datasets.

\section*{Acknowledgements}


This work was performed under the auspices of the U.S. Department of Energy by Lawrence Livermore National Laboratory under contract DE-AC52-07NA27344.
Lawrence Livermore National Security, LLC. This work was supported by the Laboratory Directed Research and Development (LDRD) program under project tracking code 23-ERD-029.
LLNL-CONF-849209.

\bibliography{main}
\bibliographystyle{icml2023}

\newpage
\appendix
\onecolumn
\section{Extended results}

\begin{figure*}[!h]
\begin{center}
\centerline{
    \includegraphics[width=0.48\columnwidth]{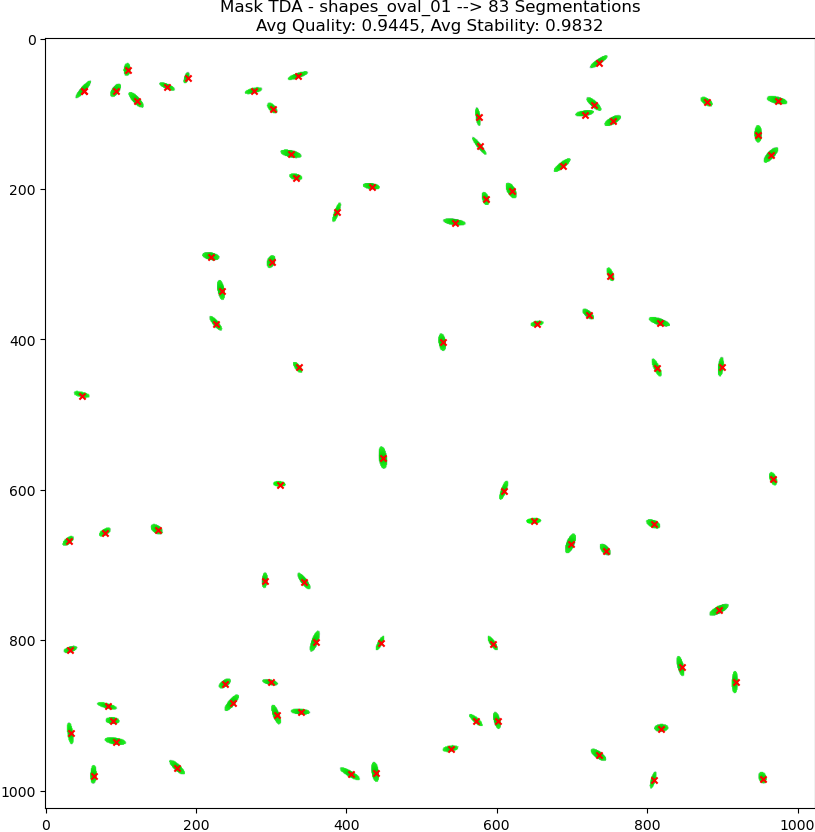}
    \includegraphics[width=0.48\columnwidth]{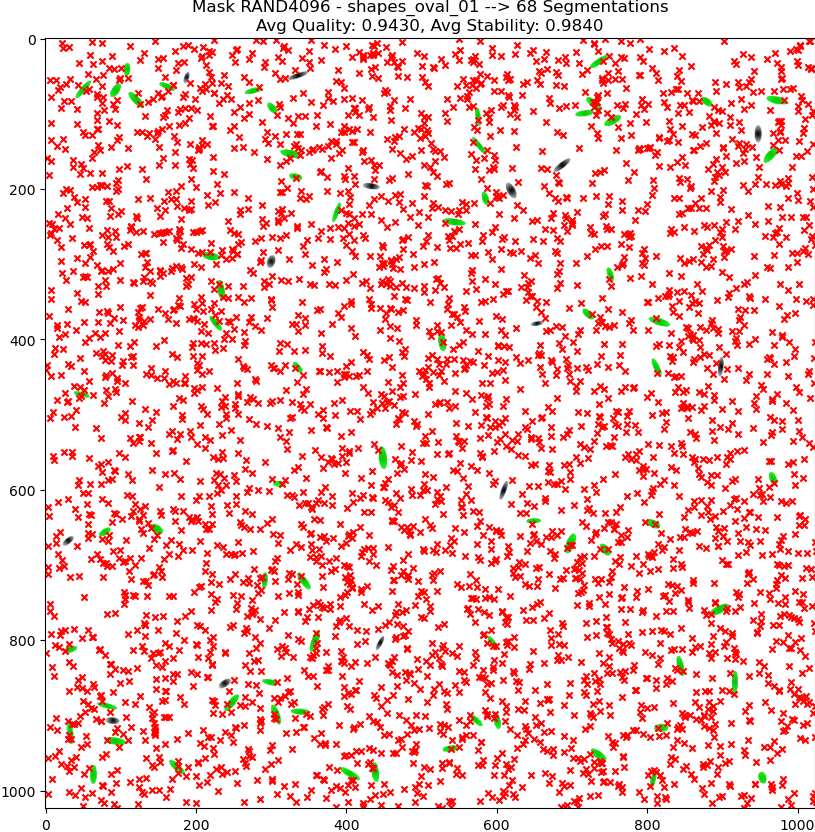}
}
\vskip 0.1in
\centerline{
    \includegraphics[width=0.48\columnwidth]{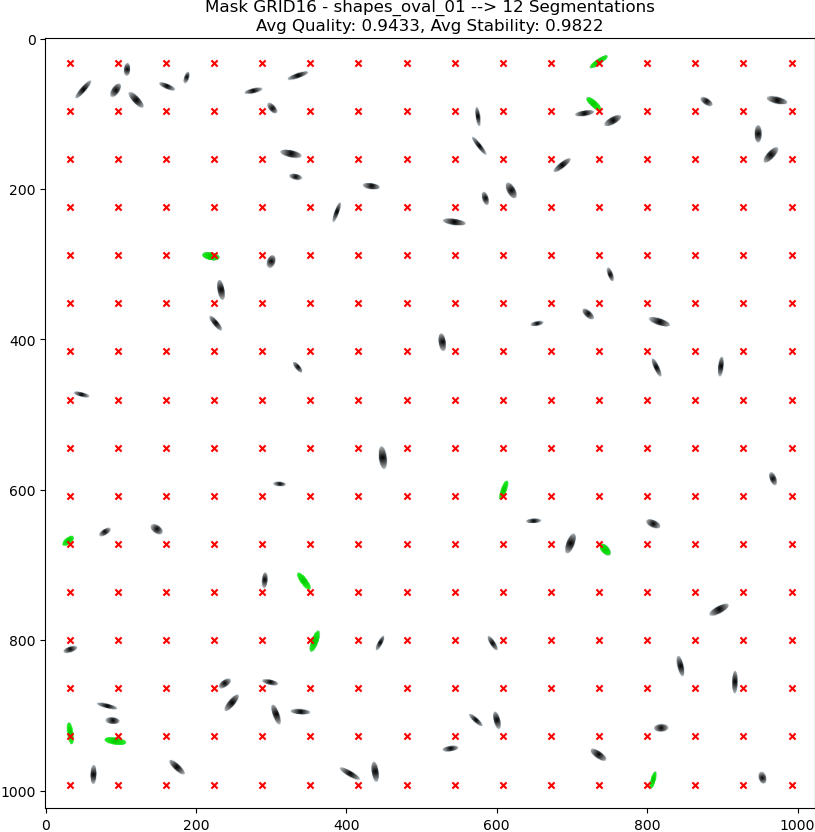}
    \includegraphics[width=0.48\columnwidth]{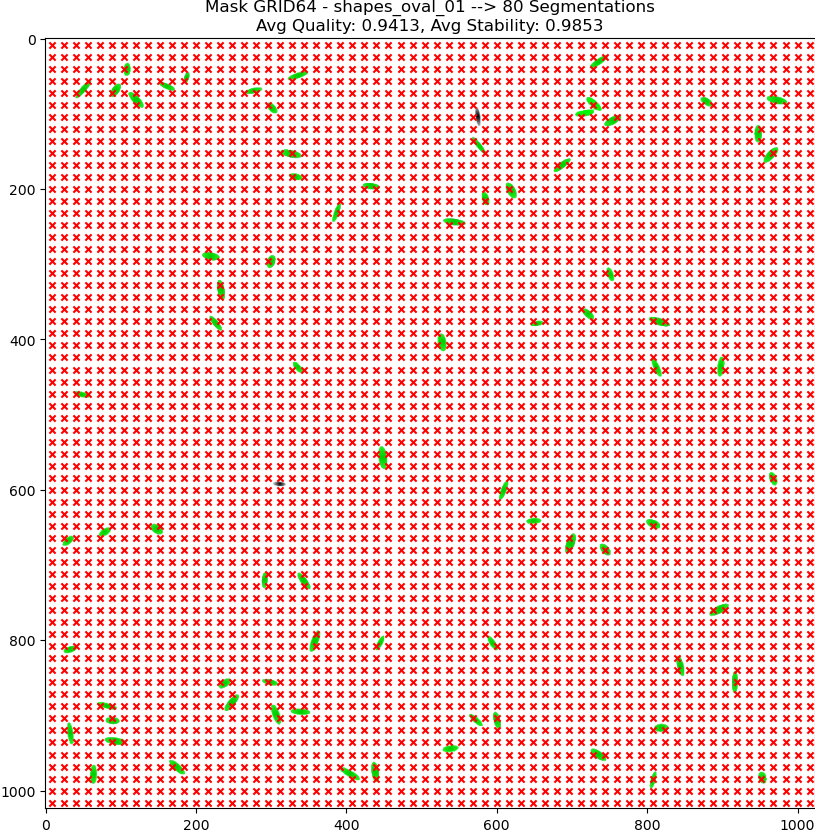}
}
\caption{Results from different point prompts for the synthetic dataset: $(top-left)$ TDA approach for optimized prompt, $(top-right)$ Very dense random point cloud with 4096 points, $(bottom-left)$ Sparse point grid with a 16 by 16 raster, and $(bottom-right)$ Very dense point grid with a 64 by 64 raster.}
\label{fig:synthetic_data_experiments}
\end{center}
\vskip -0.2in
\end{figure*}

\begin{figure*}[h]
\begin{center}
\centerline{
    \includegraphics[width=0.48\columnwidth]{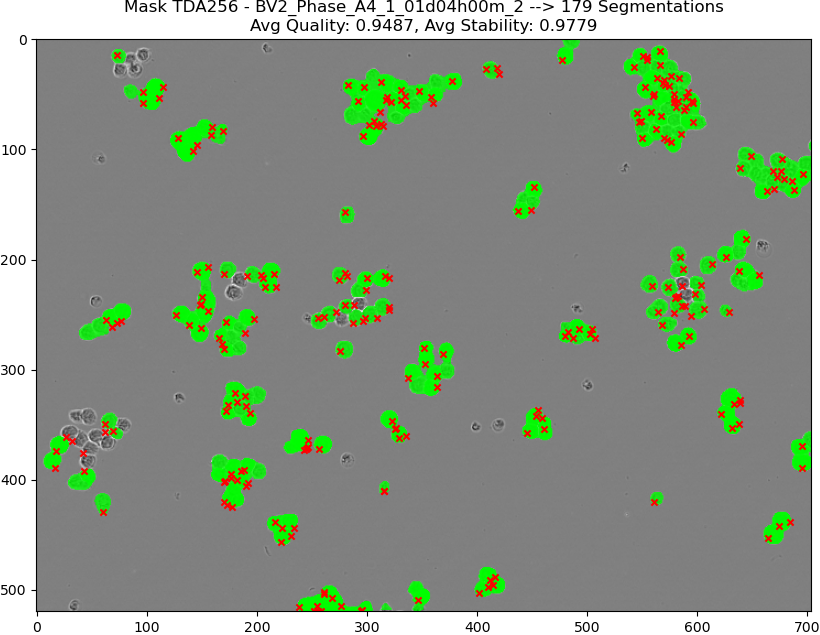}
    \includegraphics[width=0.48\columnwidth]{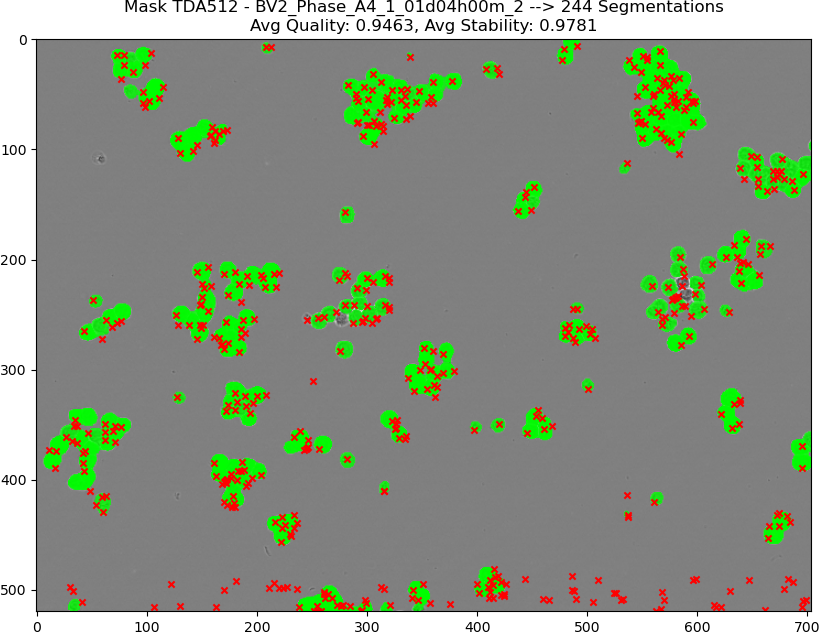}
}
\vskip 0.1in
\centerline{
    \includegraphics[width=0.48\columnwidth]{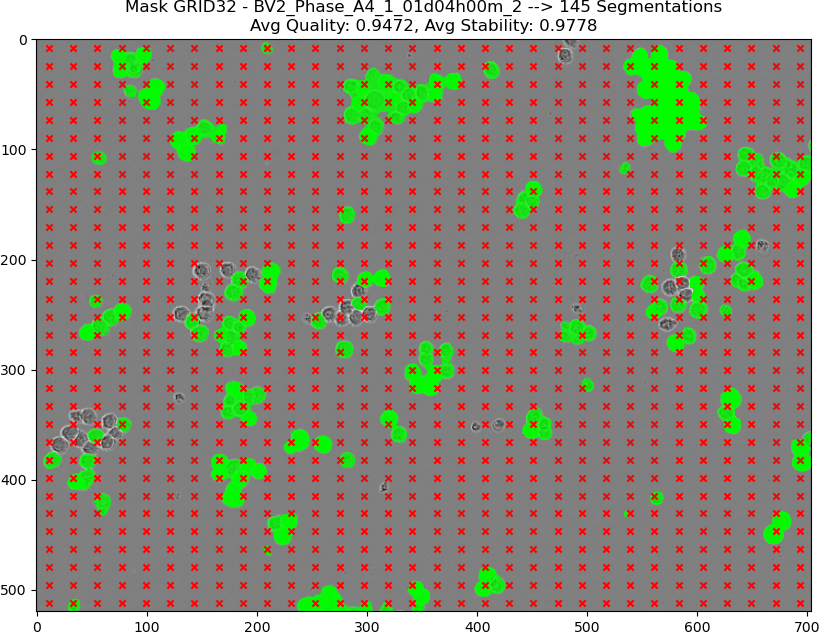}
    \includegraphics[width=0.48\columnwidth]{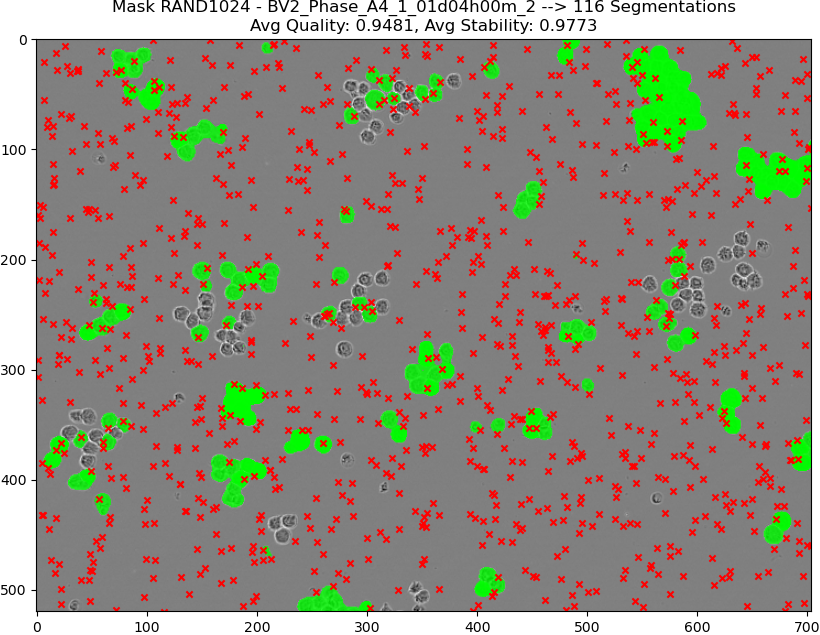}
}
\caption{Results from different point prompts for the LIVECell dataset, image $BV2\_Phase\_A4\_1\_01d04h00m\_2$: $(top-left)$ TDA approach for optimized prompt with a budget of 256 points, $(top-right)$ TDA approach for optimized prompt with a budget of 512 points, $(bottom-left)$ Point grid with a 32x32 raster, and $(bottom-right)$ Dense random point cloud with 1024 points.}
\label{fig:livecell_data_experiments}
\end{center}
\vskip -0.2in
\end{figure*}

\end{document}